\documentclass[conference]{IEEEtran}
\IEEEoverridecommandlockouts
\usepackage{cite}
\usepackage{amsmath,amssymb,amsfonts}
\usepackage{algorithmic}
\usepackage{graphicx}
\usepackage{textcomp}
\usepackage{xcolor}
\usepackage{fvextra}
\usepackage{listings}
\usepackage[T1]{fontenc}
\usepackage{lmodern} 
\usepackage{booktabs}
\usepackage[hidelinks]{hyperref}
\usepackage{subcaption}

\def\BibTeX{{\rm B\kern-.05em{\sc i\kern-.025em b}\kern-.08em
    T\kern-.1667em\lower.7ex\hbox{E}\kern-.125emX}}
\begin{document}

\title{Cybo-Waiter: A Physical Agentic Framework for Humanoid Whole-Body Locomotion–Manipulation
}

\author{
\IEEEauthorblockN{
Peng Ren\textsuperscript{1,2,*}\thanks{Equal contribution.},
Haoyang Ge\textsuperscript{3,2,*},
Chuan Qi\textsuperscript{4,2},
Cong Huang\textsuperscript{2,5},
Hong Li\textsuperscript{6},
Jiang Zhao\textsuperscript{1},
Pei Chi\textsuperscript{1,$\dagger$},
Kai Chen\textsuperscript{2,5,6,$\dagger$},
}
\IEEEauthorblockA{
\textsuperscript{1}Beihang University, Beijing, China\\
\textsuperscript{2}Zhongguancun Academy, Beijing, China\\
\textsuperscript{3}Tianjin University, Tianjin, China\\
\textsuperscript{4}University of Science and Technology of China, Hefei, China\\
\textsuperscript{5}Zhongguancun Institute of Artificial Intelligence, Beijing, China\\
\textsuperscript{6}DeepCybo, Beijing, China\\
\textsuperscript{*}Equal contribution.\\
\textsuperscript{$\dagger$}Corresponding authors: Pei Chi (peichi@buaa.edu.cn), Kai Chen (kaichen@zgci.ac.cn).
}
}

\maketitle

\begin{abstract}
Robots are increasingly expected to execute open ended natural language requests in human environments, which demands reliable long horizon execution under partial observability. This is especially challenging for humanoids because locomotion and manipulation are tightly coupled through stance, reachability, and balance. We present a humanoid agent framework that turns VLM plans into verifiable task programs and closes the loop with multi object 3D geometric supervision. A VLM planner compiles each instruction into a typed JSON sequence of subtasks with explicit predicate based preconditions and success conditions. Using SAM3 and RGB-D, we ground all task relevant entities in 3D, estimate object centroids and extents, and evaluate predicates over stable frames to obtain condition level diagnostics. The supervisor uses these diagnostics to verify subtask completion and to provide condition-level feedback for progression and replanning. We execute each subtask by coordinating humanoid locomotion and whole-body manipulation, selecting feasible motion primitives under reachability and balance constraints. Experiments on tabletop manipulation and long horizon humanoid loco manipulation tasks show improved robustness from multi object grounding, temporal stability, and recovery driven replanning.
\end{abstract}

\begin{IEEEkeywords}
Humanoid Robot, Locomotion, Reinforcement
Learning, Whole-body Control, Vision-Language-model
\end{IEEEkeywords}

\section{Introduction}

Robots are increasingly expected to execute open-ended natural-language requests in real human environments, such as
``tidy up the desk'' or ``bring me a drink''~\cite{pi_0,pi05,rt1,zhao2023chat}.
Such tasks are long-horizon and spatially extended: the robot must navigate and manipulate under clutter and occlusion while ensuring safe whole-body motion near people and furniture~\cite{rt2}.
Accordingly, many works cast instruction following as a \emph{robot agent} problem that compiles language into executable steps and revises them online.
During execution, the agent must continually decide what to do next, ground language to scene entities, and verify step completion under noisy observations,
motivating designs that couple high-level reasoning with grounded perception and reliable execution.

Under this problem setting, many language-conditioned robot systems instantiate a modular pipeline with three stages: task planning, semantic grounding, and execution evaluation.
Planning methods decompose instructions into skill sequences and may leverage feedback, e.g., feasibility-aware selection~\cite{saycan}, code-generated feedback loops~\cite{codeaspolicies}, and language-based self-feedback for replanning~\cite{innermonologue};
grounding methods map language to spatial or geometric representations for 3D-aware action generation and constraint reasoning~\cite{voxposer}.
Because long-horizon success is ultimately determined by interaction outcomes over time, platforms and benchmarks emphasize evaluation in closed-loop execution beyond offline VQA~\cite{agenticlab,antol2015vqa}.

Nevertheless, achieving robust long-horizon execution in the real world remains brittle.
Free-form model outputs can be underspecified for downstream controllers, implicit success detection is susceptible to transient perception noise, and failures are hard to attribute and recover from.
These challenges motivate explicit verification and replanning, such as logic-based plan checking~\cite{verifyllm} and scene-perception-driven proactive replanning~\cite{scenegraphreplan,replanvlm}.
However, most existing frameworks are validated on mobile manipulators or wheeled platforms and do not fully account for the tighter coupling and safety constraints of humanoid whole-body execution.

This limitation becomes even more critical for humanoid agents, where whole-body constraints tightly couple locomotion and manipulation through stance, reachability, and balance \cite{beyondmimic,Deepmimic,iwalker,ze2025generalizable}.
Consequently, small distance and pose errors can cascade into failures during subsequent locomotion and manipulation.
Recent humanoid systems can pursue two directions:
(i) hierarchical agent stacks that improve online efficiency via learned intermediate decision layers and modular skills \cite{being0},
as well as VLM-based monitoring for skill sequencing on real humanoids~\cite{hierarchicalhumanoidvlp}; and
(ii) learning-heavy vision-language-action policies trained at scale for large-space whole-body loco-manipulation, studied in either real-robot or large-scale simulation settings~\cite{wholebodyvla,humanoidverse}.
While highly promising, these approaches typically emphasize efficiency or broad generalization.
In contrast, a key requirement for a dependable humanoid agent remains under-explored:
a \emph{verifiable and diagnosable} execution loop that grounds all task-relevant entities in actionable 3D geometry, evaluates explicit success conditions with temporal stability under noisy perception, and
 triggers targeted recovery when conditions are violated, helping to avoid premature termination and unsafe motions.

To address this, we propose a training-light humanoid agent framework that turns open-ended VLM plans into \emph{verifiable task programs} 
and closes the loop with \emph{multi-object 3D geometric supervision}.
Given an instruction, a VLM-based planner compiles it into a sequence of structured subtasks in a typed JSON schema, 
where each subtask specifies task-conditioned entity descriptions
and explicit preconditions/success conditions as predicate assertions with temporal stability.
We then ground \emph{all task-relevant entities}, including the manipulation target, the destination region, and any objects referenced by predicates, in 3D using SAM3 segmentation and RGB-D observations, recovering object-centric geometric states and pairwise relations \cite{sam3}.
A geometry-grounded supervisor evaluates predicates over stable frames, produces condition-level diagnostics, and triggers targeted recovery, such as re-grounding, re-approach adjustment, or replanning.
Finally, these supervision signals are integrated with whole-body execution by gating locomotion and MPC-based upper-body manipulation 
through task-conditioned feasibility and safety constraints, enabling robust long-horizon humanoid loco-manipulation.

In this paper, our main contributions are:
\begin{itemize}
  \item A structured and verifiable task interface that compiles VLM outputs into typed subtasks with explicit predicate-based preconditions and success conditions.
    \item Task-conditioned \emph{multi-object} 3D grounding using SAM3+RGB-D to recover object-centric geometric states and relations for verification.
    \item A geometry-grounded supervisor with temporal stability and condition-level diagnostics that enables targeted recovery and feedback-driven replanning.
    \item An integrated humanoid execution layer that bridges geometric supervision to whole-body control for long-horizon loco-manipulation.
\end{itemize}

\section{Related Work}

\subsection{Language-conditioned Robot Agents}
Recent language-conditioned robot agents leverage LLM/VLMs to decompose open-ended instructions into modular skill sequences and execute them with varying degrees of online feedback, e.g., feasibility-aware skill selection~\cite{saycan}, program/code-based policies with embedded feedback loops~\cite{codeaspolicies}, zero-shot language planning~\cite{huang2022language}, and self-reflective replanning via language feedback~\cite{innermonologue}, with benchmarks further emphasizing that reliable evaluation should go beyond offline VQA and include closed-loop execution~\cite{agenticlab,yao2022react}.

\subsection{Grounding and Geometric Verification for Execution}
Complementary to high-level planning, prior work grounds language into spatial or object-centric representations for action generation and verification, including 3D-aware grounding and affordance reasoning~\cite{voxposer,cliport,peract} as well as explicit verification and replanning mechanisms such as logic-based plan checking~\cite{verifyllm} and scene-graph/perception-driven proactive replanning~\cite{scenegraphreplan,replanvlm}, but these approaches often provide limited condition-level diagnostics tied to multi-object 3D geometry.

\subsection{Humanoid Agents and Whole-Body Loco-Manipulation}
Humanoid agents pose additional challenges due to whole-body coupling between locomotion, balance, and manipulation, and recent systems pursue either hierarchical stacks that improve online efficiency via connector-style intermediate decision layers and modular skills~\cite{being0,Leverb,wang2024autonomous} or VLM-based monitoring for skill sequencing on real humanoids~\cite{hierarchicalhumanoidvlp}, or learning-heavy vision-language-action policies trained at scale for large-space whole-body loco-manipulation~\cite{wholebodyvla,humanoidverse,radosavovic2024real}.

\section{Methods}
\begin{figure*}[t]
  \centering
  \includegraphics[width=0.9\textwidth]{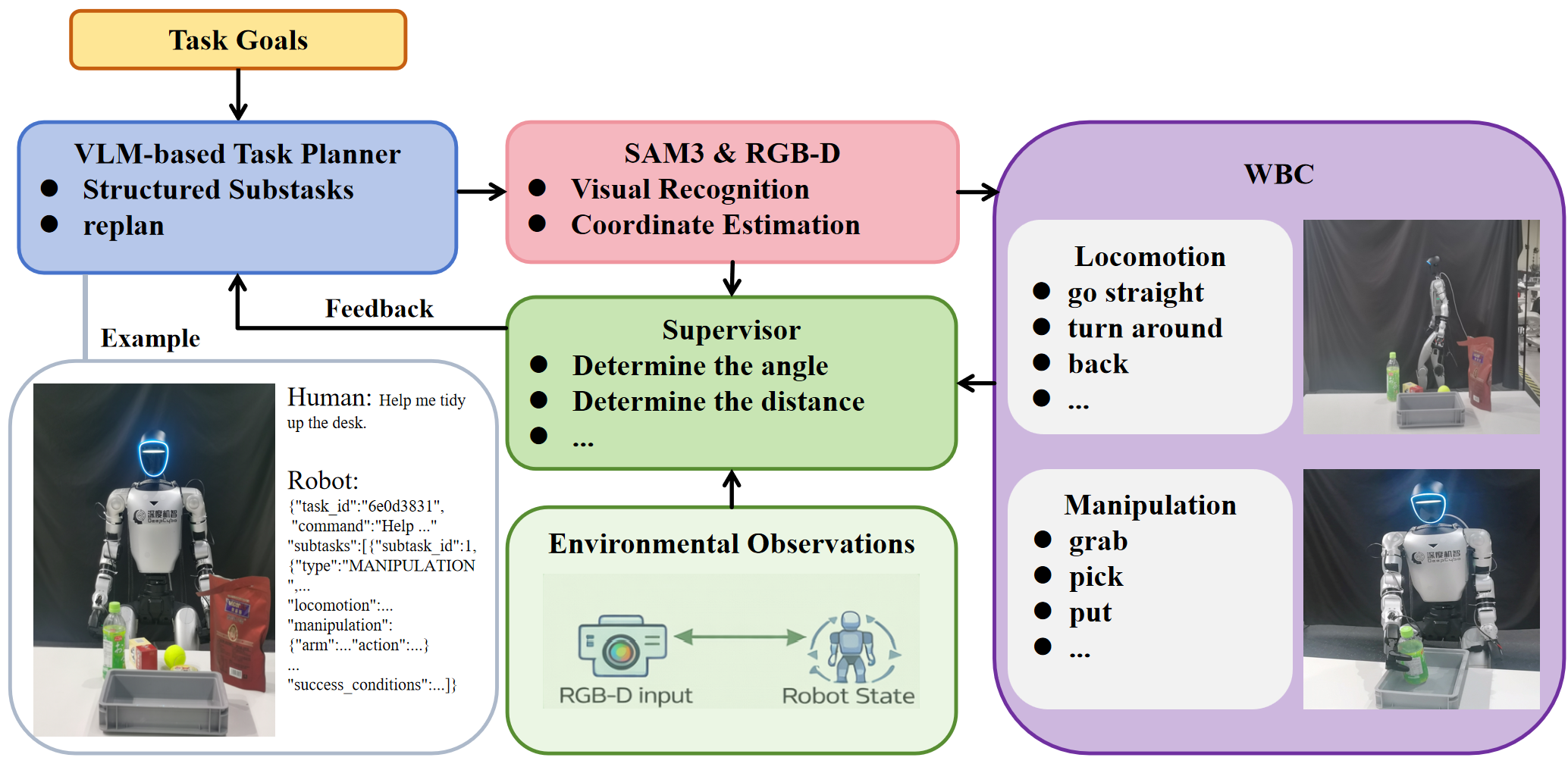}
  \caption{System overview of the humanoid agent.}
  \label{fig:system}
\end{figure*}

We address long-horizon humanoid loco-manipulation from a natural language instruction $I$ using RGB-D observations and robot state in a closed loop.
Our system, as shown in Fig.~\ref{fig:system}, comprises structured VLM task decomposition, task conditioned multi object Tracking, predicate based supervision and targeted recovery, feedback driven replanning, and integrated whole body execution for locomotion and MPC-based manipulation.

\subsection{Structured VLM Task Decomposition}\label{AA}

We decouple semantic planning from perception and execution through a structured subtask interface.
Given a high level command $I$ and observation $o_t$, a VLM-based planner generates a task plan
$\mathcal{P}=\langle task\_id, I, \{\tau_i\}_{i=1}^{N}\rangle$.
We compile form VLM outputs into a typed JSON.
\begin{equation}
\begin{aligned}
\tau_i=&\langle \texttt{subtask\_id},~\texttt{type},~\texttt{target},~\texttt{destination},\\
&~\texttt{manipulation},~\texttt{locomotion},~\texttt{preconditions},\\
&~\texttt{success\_conditions},~\texttt{failure\_handlers},\\
&~\texttt{timeout\_sec},~\texttt{max\_retry}\rangle.
\end{aligned}
\end{equation}
Here, type identifies the subtask family. 
The field \texttt{target} specifies the object to be grounded through a symbolic handle (\texttt{ref}) and a language descriptor (\texttt{phrase}), optionally augmented with \texttt{category}, \texttt{attributes}, and \texttt{relations}. 
We use \texttt{destination} to describe the intended goal region or reference object in natural language. 
The \texttt{manipulation} block instantiates an action primitive together with actuation settings such as arm selection and whether to enable upper-body MPC. 
Finally, \texttt{preconditions} and \texttt{success\_conditions} are represented as predicate assertions with temporal stability specified by \texttt{stable\_frames}, which enables deterministic monitoring and closed-loop replanning. An example is provided in Appendix \ref{appendix}.

We detail how task-conditioned specifications are grounded into segmentation and 3D geometry, and how predicates are evaluated over stable frames in next subsection.

\subsection{Conditional Segmentation and Geometric Coordinate Estimation}\label{seg}
This module grounds the task-conditioned plan from Sec.~\ref{AA} into \emph{multi-object} 3D geometric estimates.
Given an RGB-D observation $o_t$ and the active subtask $\tau_i$, we identify a set of task-relevant entities
\begin{equation}
    \begin{aligned}
        \mathcal{E}_i =& \{\texttt{target}\} \cup \{\texttt{destination}\} \cup \{\texttt{relation\;references}\}  \\ & \cup \{\texttt{predicate\;references}\},
    \end{aligned}
\end{equation}
where \texttt{relation\;references} are objects mentioned in \texttt{target.relations}, and \texttt{predicate\;references} are objects appearing in
\texttt{preconditions}/\texttt{success\_conditions} arguments, e.g., the \texttt{support} object in \texttt{SUPPORTED\_BY}.
This ensures that both the manipulation target and goal/reference entities are segmented and localized.

\paragraph{Task-conditioned multi-object segmentation.}
For each entity $e\in\mathcal{E}_i$, we query SAM3 to obtain a set of candidate masks $\mathcal{M}_e=\{m_{e,k}\}$ \cite{sam3}.
The query is conditioned on the plan fields: \texttt{target.phrase} prompts segmentation of the manipulated object,
\texttt{destination.phrase} prompts segmentation of the goal container/region, and referenced entities are prompted by their associated phrases or predicate roles.
Optional attributes and relations refine the query to resolve ambiguity.
When multiple instances are detected, we maintain temporal consistency for each entity by favoring candidates that remain stable across time.
For entities without an explicit \texttt{ref} in the plan, we create an internal identifier and maintain a mapping in the workspace state.

\paragraph{Lifting masks to 3D geometric states.}
For each candidate mask $m_{e,k}$, we back-project masked depth pixels into a point cloud $P_{e,k}$ using calibrated RGB-D intrinsics,
and estimate a compact geometric state:
(i) centroid $c_{e,k}$ in the world frame,
(ii) spatial extent $b_{e,k}$,
and (iii) confidence $q_{e,k}$ combining segmentation confidence with depth validity and point support.
For subtasks requiring orientation cues, we additionally estimate an orientation descriptor from $P_{e,k}$,
such as a dominant surface normal or a principal axis via PCA.
To improve robustness under sensor noise and partial observations, we apply temporal filtering and reject outliers with inconsistent depth,
insufficient inlier support, or abrupt geometric changes.

\paragraph{Candidate selection and object-centric representation}
For each entity $e\in\mathcal{E}_i$, we select the best instance by jointly scoring semantic consistency, relational consistency,
geometric feasibility, and temporal stability:
\begin{equation}
    \begin{aligned}
        S_e(k)=&w_s S_{\text{sem}}(e,k)+w_r S_{\text{rel}}(e,k)
        +w_g S_{\text{geo}}(e,k)\\&+w_t S_{\text{temp}}(e,k),
\\
k_e^\star&=\arg\max_k S_e(k).
    \end{aligned}
\end{equation}
The selected target instance is packaged as the handle
$h_i=\langle \texttt{ref},~m_{\texttt{target},k^\star_{\texttt{target}}},~c_{\texttt{target},k^\star_{\texttt{target}}},~b_{\texttt{target},k^\star_{\texttt{target}}},~q_{\texttt{target},k^\star_{\texttt{target}}}\rangle$,
while all entities, including \texttt{destination} and predicate references, are stored in an object-centric workspace state $\mathcal{W}_t$.
From $\mathcal{W}_t$, we derive pairwise geometric relations such as object--container distance, object--support height difference,
end-effector--object distance, and relative alignment, which are consumed by the supervisor to evaluate plan predicates and trigger recovery or replanning when needed.

\subsection{Supervisor: Task Monitoring with RGB-D and SAM3}\label{sup}

\paragraph{Role and inputs}
The supervisor verifies task progress and completion in a closed loop.
At each time step $t$, it consumes the active subtask $\tau_i$ from the plan (Sec.~\ref{AA}),
the target handle $h_i=\langle \texttt{ref}, m, c, b, q\rangle$, and the object-centric workspace state $\mathcal{W}_t$
constructed in Sec.~\ref{seg}.
It outputs a discrete subtask status (\texttt{in\_progress}, \texttt{done}, \texttt{blocked}, \texttt{failed})
together with an \texttt{uncertain} flag, condition-wise flags, and continuous diagnostic metrics to support recovery and replanning.

\paragraph{Predicate representation}
For each active subtask $\tau_i$, the plan specifies a set of preconditions $\mathcal{P}_i$
and success conditions $\mathcal{G}_i$.
Each condition is a parameterized predicate
\(
p=\langle k, a, \mathrm{op}, v, n\rangle
\),
where $k$ is a predicate key such as \texttt{VISIBLE}, \texttt{SUPPORTED}\_BY, \texttt{SUPPORTED\_BY}\texttt{INSIDE},
$a$ denotes grounded arguments instantiated from the plan,
$\mathrm{op}$ is a comparator, $v$ is a target value, and $n$ is the required stability length
given by \texttt{stable\_frames}.
Let $\mathcal{W}_t$ be the object-centric geometric workspace state at time $t$,
which contains per-object estimates, including mask, centroid, extent, confidence, and optional orientation descriptors, as well as derived pairwise relations.
We evaluate predicates via a monitoring function $\phi(k,a;\mathcal{W}_t)\in\mathbb{R}$ that returns a scalar geometric measure parameterized by $a$.
We treat safety/feasibility checks as invariant preconditions that must remain satisfied throughout execution.

\paragraph{Evaluation, temporal stability, and subtask status.}
For a predicate $p=\langle k,a,\mathrm{op},v,n\rangle$, we compute a scalar measure
$\phi(k,a;\mathcal{W}_t)\in\mathbb{R}$ from the geometric workspace state $\mathcal{W}_t$,
and convert it into an instantaneous boolean value:
\begin{equation}
\mathrm{eval}(p,t)=\mathbf{1}\!\left(\phi(k,a;\mathcal{W}_t)\ \mathrm{op}\ v\right)\in\{0,1\},
\end{equation}
where $\mathbf{1}(\cdot)$ is an indicator function that equals 1 if its argument is true and 0 otherwise.
To suppress transient noise, we require the predicate to hold for $n$ consecutive frames:
\begin{equation}
\mathrm{sat}(p,t)=\prod_{u=t-n+1}^{t}\mathrm{eval}(p,u),
\end{equation}
which equals 1 iff $\mathrm{eval}(p,u)=1$ for all $u\in[t-n+1,t]$.

We define readiness and completion as conjunctions over predicate sets:
\begin{equation}
\begin{aligned}
\mathrm{Ready}(\tau_i,t)&=\bigwedge_{p\in\mathcal{P}_i}\mathrm{sat}(p,t),\\
\mathrm{Done}(\tau_i,t)&=\bigwedge_{g\in\mathcal{G}_i}\mathrm{sat}(g,t),
\end{aligned}
\end{equation}
where $\wedge$ denotes logical AND.
If $\mathrm{Ready}(\tau_i,t)=0$, the supervisor marks the subtask as \texttt{blocked} and reports the failing predicates.
If $\mathrm{Done}(\tau_i,t)=0$ for longer than \texttt{timeout\_sec}, or if safety/feasibility predicates are violated,
the supervisor marks \texttt{failed} and triggers recovery or replanning; otherwise it remains \texttt{in\_progress}.

\paragraph{Diagnostics and corrective feedback}
Beyond discrete subtask status, the supervisor returns a diagnostic vector $\mathbf{d}_t$ computed from $\mathcal{W}_t$,
including continuous measures such as object--support height gap, relative distance, alignment angle, and confidence/stability scores.
These metrics localize which constraints/predicates fail and quantify the magnitude of error, enabling targeted corrections.
When estimates are unreliable, the supervisor requests re-observation or re-grounding before proceeding.
The diagnostic vector is forwarded to downstream modules as structured signals for corrective motion and replanning.

\paragraph{VLM-assisted semantic verification}
In addition to geometric predicates, we optionally invoke a VLM as a semantic verifier when the supervisor raises the \texttt{uncertain} flag,
or when completion depends on non-geometric cues.
Given the active subtask $\tau_i$ and a compact visual context (RGB image, optional masks, and key geometric diagnostics),
the VLM outputs a boolean judgment and a short rationale about whether the subtask is completed.
We use this signal as \emph{auxiliary} feedback rather than a primary success criterion:
geometric verification takes precedence whenever predicates are reliably evaluable from $\mathcal{W}_t$,
and VLM verification is consulted only when geometric evidence is missing, ambiguous, or low-confidence.
If the two signals disagree, the system triggers re-observation or re-grounding before declaring success.

\subsection{Feedback-Driven Replanning for Robust Task Execution}\label{SCM}

\paragraph{Replanning triggers}
Replanning is triggered by structured feedback from the supervisor.
Concretely, we initiate replanning when (i) a subtask becomes \texttt{blocked} due to unsatisfied preconditions,
(ii) a subtask is marked \texttt{failed},
or  (iii) the supervisor raises an \texttt{uncertain} flag due to low-confidence or inconsistent geometric estimates.

\paragraph{Structured feedback}
At runtime, the supervisor returns (1) the discrete subtask status, (2) the set of failing predicate instances and their grounded arguments,
and (3) continuous diagnostics $\mathbf{d}_t$.
Conditioned on the subtask phase, we define the failing predicate set as
\[
\mathcal{P}^{\text{fail}}_t=
\begin{cases}
\{p\in \mathcal{P}_i ~|~ \mathrm{sat}(p,t)=0\}, & \texttt{status}=\texttt{blocked},\\
\{g\in \mathcal{G}_i ~|~ \mathrm{sat}(g,t)=0\}, & \texttt{status}=\texttt{failed},
\end{cases}
\]
and form a compact feedback record
$\mathcal{F}_t=\langle \tau_i,\ \texttt{status},\ \mathcal{P}^{\text{fail}}_t,\ \mathbf{d}_t\rangle$,
which is consumed by the planner to decide whether to re-observe, re-ground, retry with modified parameters, or replan the remaining subtasks.
For \texttt{in\_progress}, we primarily use $\mathbf{d}_t$ to diagnose progress and select corrective actions.

\paragraph{Recovery actions}
We employ a small set of reusable recovery operators conditioned on the failure type:
(i) \emph{Re-observation}: adjust viewpoint or pause to acquire a more reliable RGB-D frame,
(ii) \emph{Re-grounding}: re-run conditional segmentation and 3D estimation with updated prompts/relations,
(iii) \emph{Skill/parameter adaptation}: switch the manipulation primitive or refine target tolerances based on $\mathbf{d}_t$, micro-adjust end-effector pose when alignment error is small,
and (iv) \emph{Task-level revision}: modify the remaining plan by inserting corrective subtasks.
These operators can be specified in \texttt{failure\_handlers} and executed before calling the VLM planner for full replanning.

\paragraph{Retry budget and plan update}
Each subtask carries an execution budget via \texttt{timeout\_sec} and \texttt{max\_retry}.
If the retry budget is not exhausted, we re-execute $\tau_i$ after applying a recovery operator;
otherwise, we request the planner to generate an updated plan $\mathcal{P}'$ conditioned on $\mathcal{F}_t$,
while preserving completed subtasks and updating symbolic references for consistent grounding.

\subsection{Integrated Execution Framework for Robust Humanoid Task Execution}\label{exec}

\paragraph{Execution interface}
The execution layer instantiates plan-level subtasks $\tau_i$ into low-level humanoid controllers.
It takes as input the active subtask schema  together with the grounded target handle $h_i$,
and dispatches either locomotion or manipulation skills depending on \texttt{type}.
Crucially, the layer follows the plan fields \texttt{locomotion} and \texttt{manipulation} as an explicit control interface, and returns execution status and telemetry to the supervisor for monitoring and replanning.

\paragraph{Locomotion as a gait-conditioned RL skill library}
For \texttt{LOCOMOTION} subtasks, we use a single reinforcement-learned locomotion policy as a stand-alone skill.
The policy is \emph{gait-conditioned}: at runtime the agent selects a discrete gait identifier $\texttt{gait\_id}$, and the policy produces
lower-body commands for robust balance and stepping while tracking desired base velocity and yaw-rate commands.
We train this policy with a multi-clip locomotion dataset consisting of 12 motion clips, covering forward/backward walking, lateral stepping,
running, and in-place turning.

To expose a compact action interface to the planner/supervisor, we map the 12 clip-level gaits into 7 locomotion primitives:
\emph{forward walk} ($\texttt{gait\_id}\in\{0,1\}$),
\emph{backward walk} ($\texttt{gait\_id}\in\{2,3\}$),
\emph{carry-and-walk} ($\texttt{gait\_id}\in\{4,5\}$),
\emph{run} ($\texttt{gait\_id}\in\{6,7\}$),
\emph{side-walk} ($\texttt{gait\_id}\in\{8,9\}$),
\emph{turn-left} ($\texttt{gait\_id}=10$),
and \emph{turn-right} ($\texttt{gait\_id}=11$).
For primitives with multiple available clips, we either select a canonical clip or sample within the clip group, which improves coverage and reduces overfitting to a single clip.
This gait-conditioned design provides a unified locomotion skill library for long-horizon execution and recovery, enabling task-dependent repositioning without hand-coded motion scripts.

The policy is trained in simulation to track commanded base velocities and yaw rates while maintaining balance and contact stability.
We condition the policy on $\texttt{gait\_id}$ and train on a mixture of motion clips with domain randomization to improve robustness.

\paragraph{Manipulation with decoupled upper-body control.}
For \texttt{MANIPULATION} subtasks, the execution layer performs arm motion while maintaining lower-body stability.
The arm and action are specified by the plan, and the target handle $h_i$ provides the geometric reference for motion generation.
When \texttt{use\_upper\_body\_mpc} is enabled, we solve an upper-body MPC problem to produce constraint-aware end-effector trajectories.
Meanwhile, the lower body is controlled independently to preserve balance and support stance, allowing manipulation to proceed
without destabilizing the humanoid.

\paragraph{Cross-skill coordination and transitions.}
Long-horizon tasks typically require alternating between repositioning and manipulation.
We implement a lightweight coordinator that switches between the RL locomotion skill and the upper-body manipulation controller
based on subtask type and supervisor feedback.
In particular, if geometric preconditions are not satisfied, the coordinator triggers a locomotion subskill to reposition
until the supervisor marks the subtask \texttt{ready}; if manipulation tracking errors increase or safety predicates are violated,
the coordinator pauses arm motion and requests recovery actions.
This decoupled design enables robust locomotion-manipulation execution while preserving a clean agent-facing interface
defined by the JSON plan.

\section{EXPERIMENTS}

\begin{figure*}[t]
  \centering
  \begin{subfigure}[t]{0.24\textwidth}
    \centering
    \includegraphics[width=\linewidth,height=5cm]{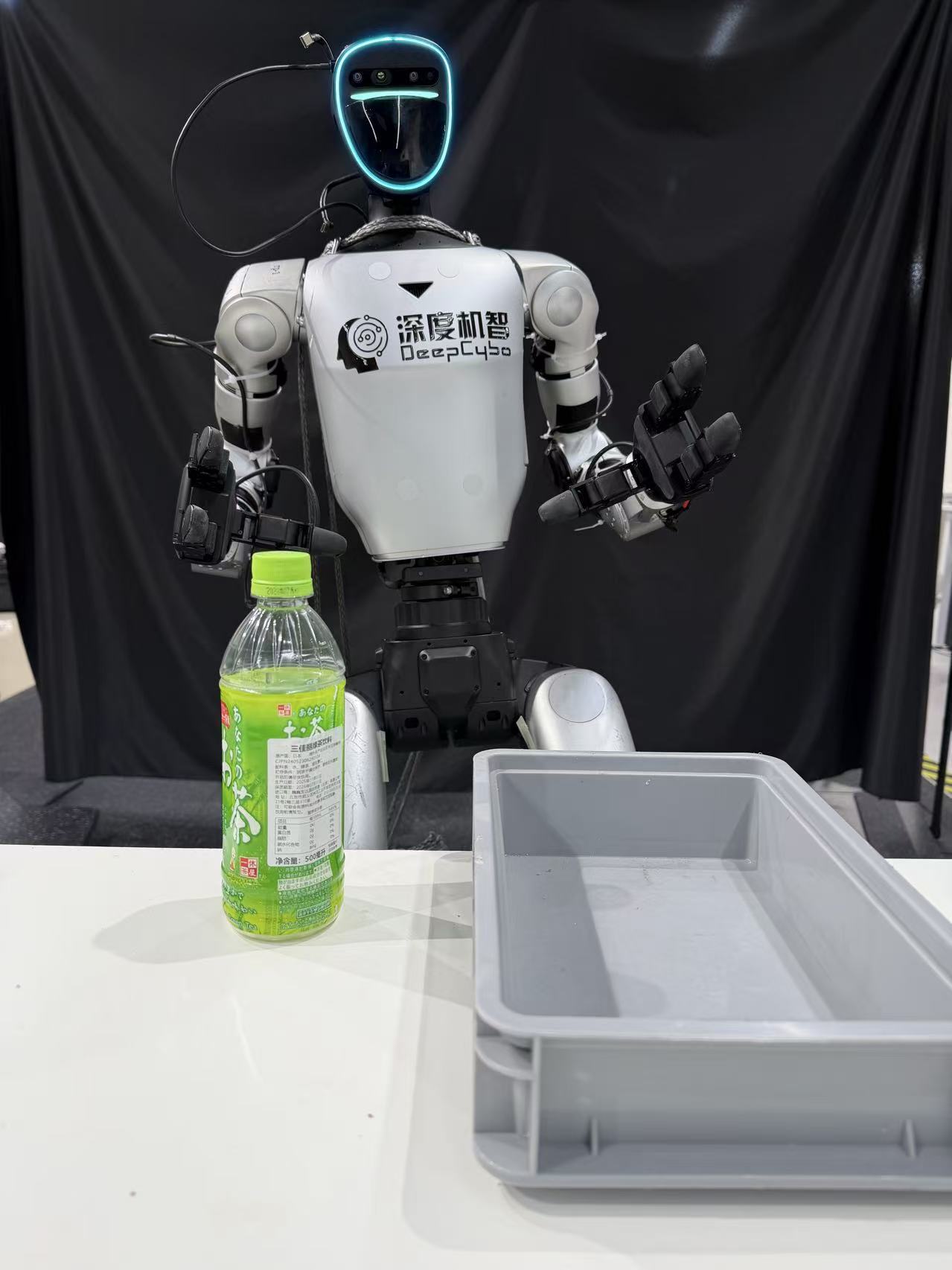}
    \caption{Initial scene with the bottle and tray.}
  \end{subfigure}\hfill
  \begin{subfigure}[t]{0.24\textwidth}
    \centering
    \includegraphics[width=\linewidth,height=5cm]{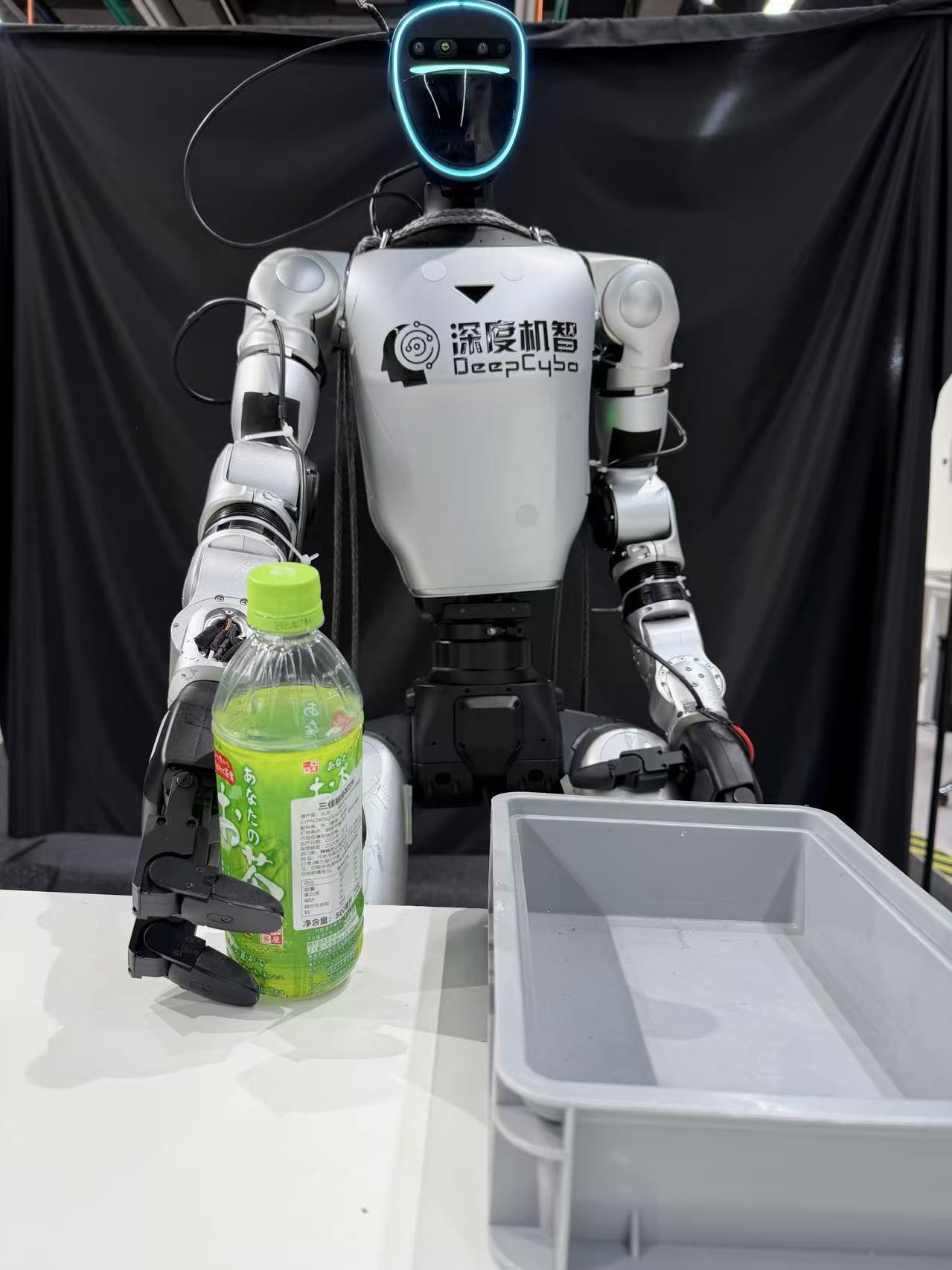}
    \caption{Reaching and grasping the bottle.}
  \end{subfigure}\hfill
  \begin{subfigure}[t]{0.24\textwidth}
    \centering
    \includegraphics[width=\linewidth,height=5cm]{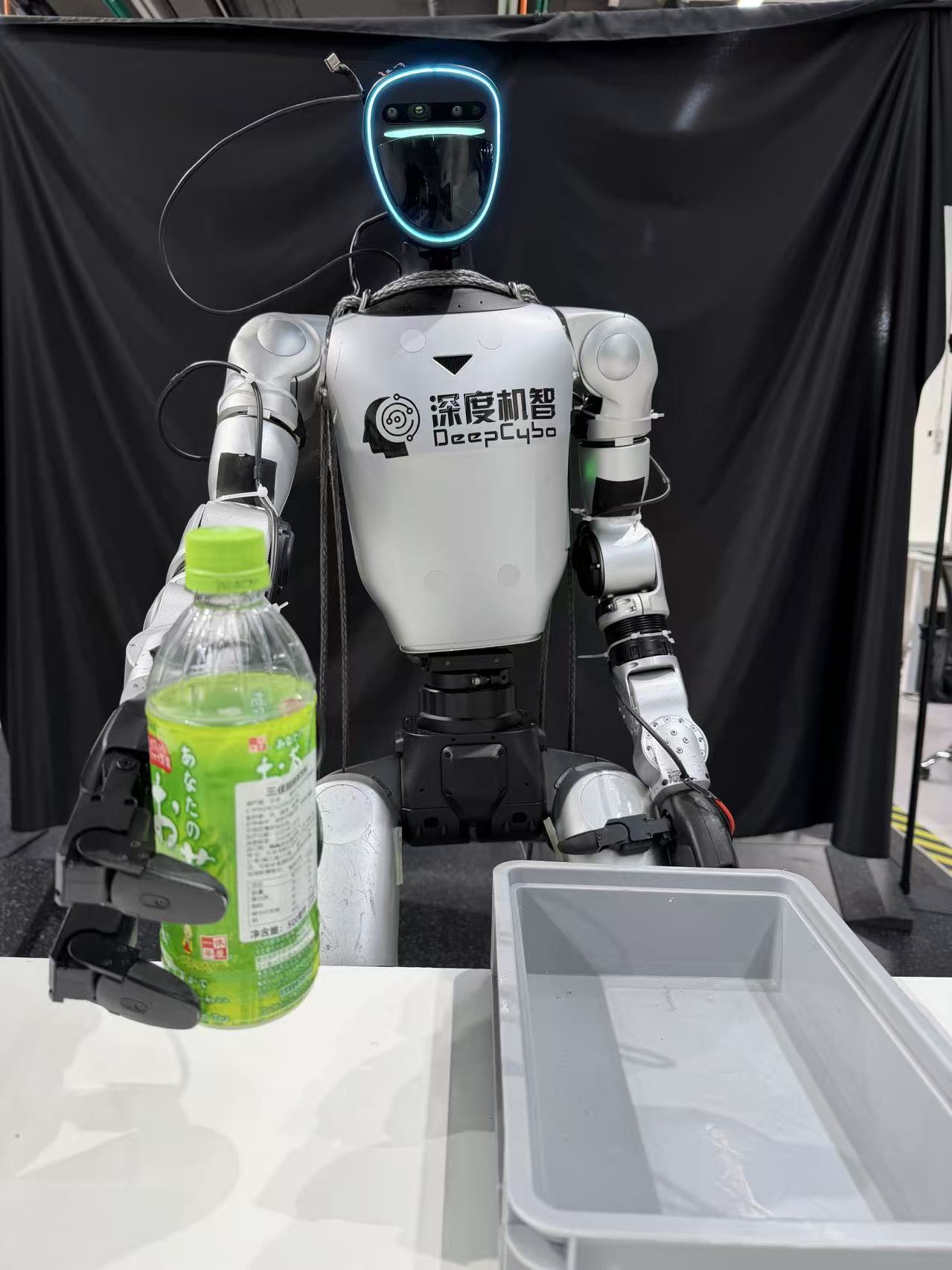}
    \caption{Lifting and transporting the bottle toward the tray.}
  \end{subfigure}\hfill
  \begin{subfigure}[t]{0.24\textwidth}
    \centering
    \includegraphics[width=\linewidth,height=5cm]{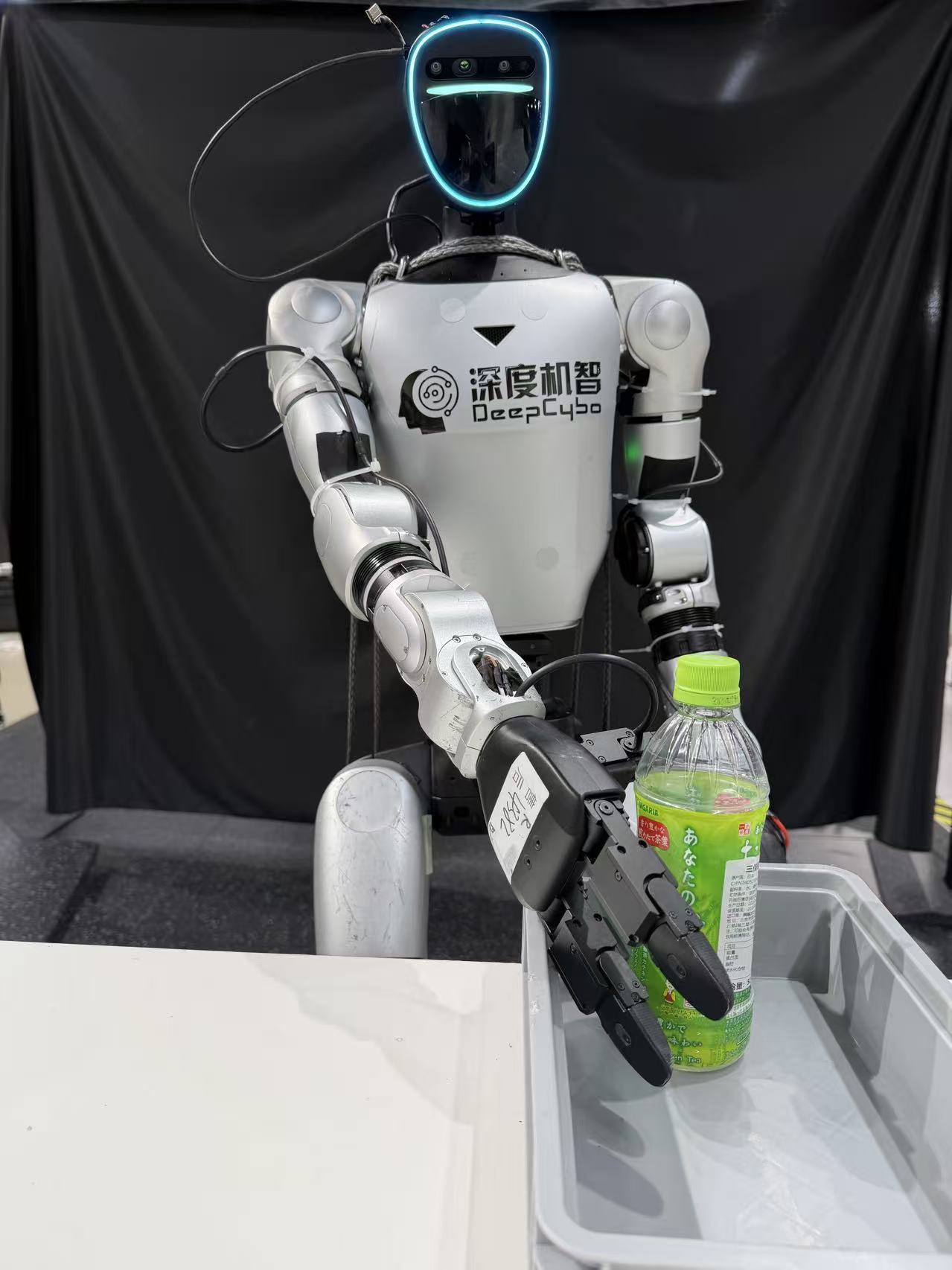}
    \caption{Placing the bottle into the tray.}
  \end{subfigure}
  \caption{Partial “Tidy-the-desk” subtasks: bottle pick-and-place into a tray.}
  \label{fig:51}
\end{figure*}

\begin{figure*}[t]
  \centering
  \begin{subfigure}[t]{0.24\textwidth}
    \centering
    \includegraphics[width=\linewidth,height=5cm]{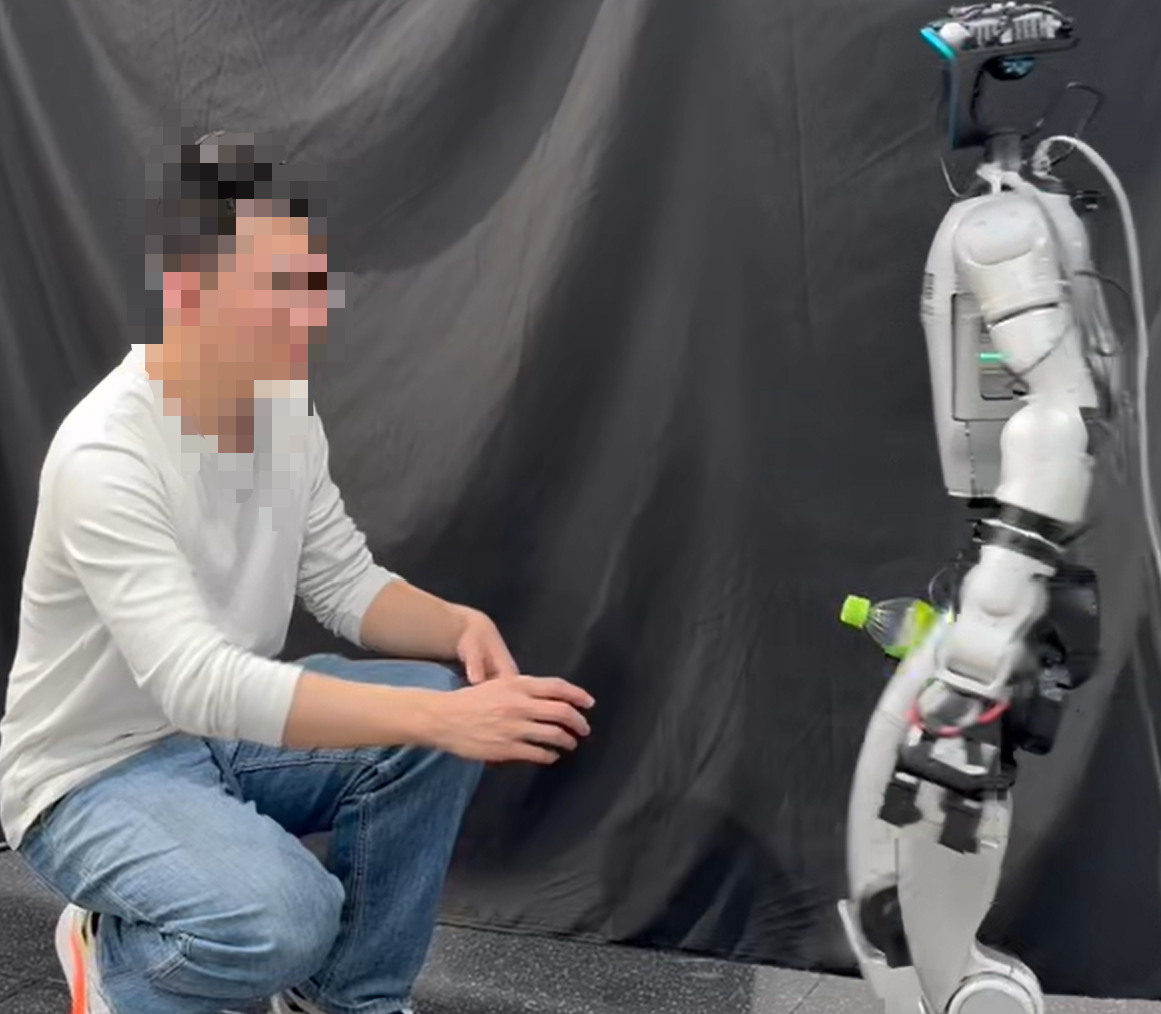}
    \caption{Searching for the user.}
  \end{subfigure}\hfill
  \begin{subfigure}[t]{0.24\textwidth}
    \centering
    \includegraphics[width=\linewidth,height=5cm]{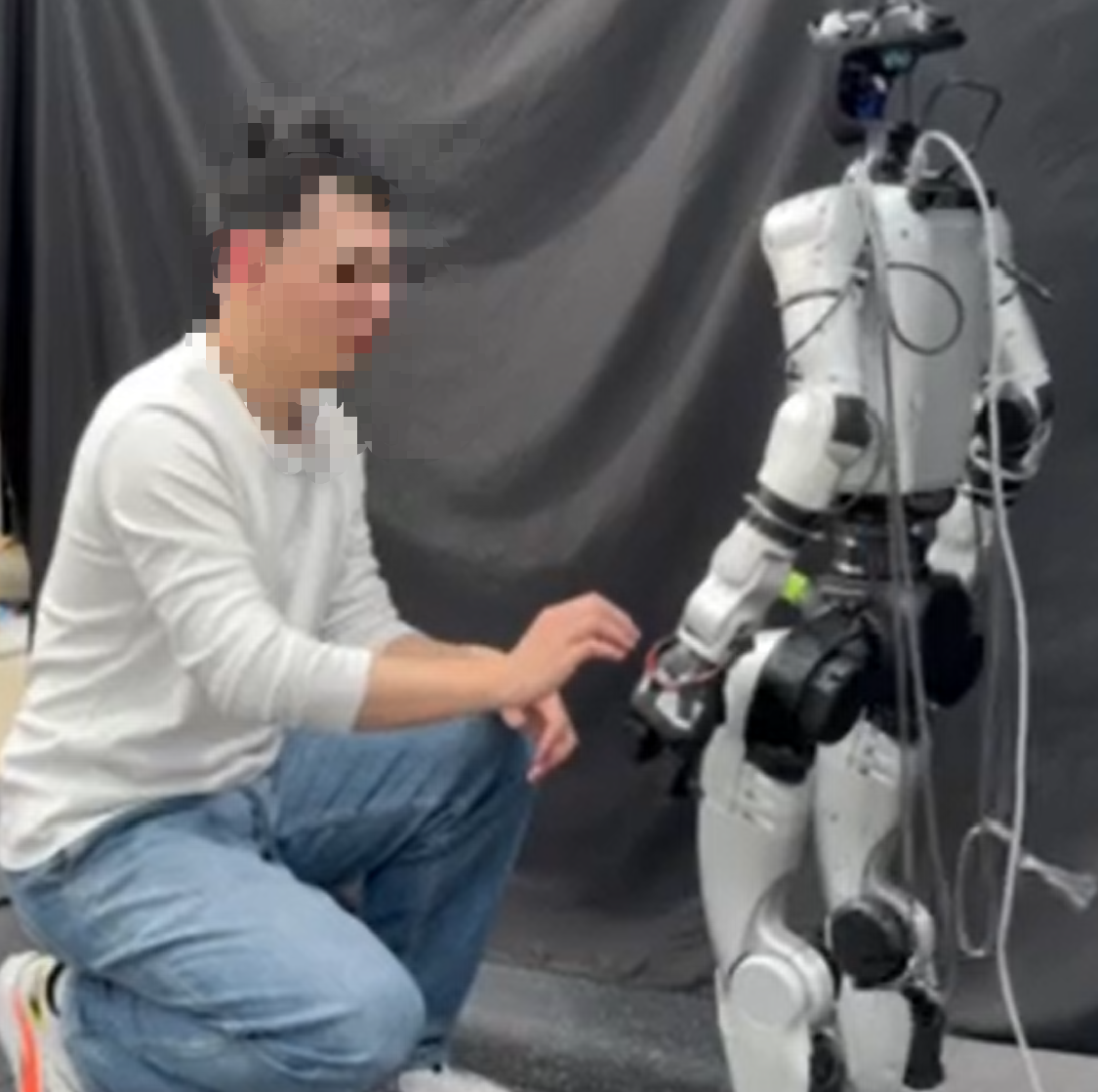}
    \caption{Approaching the user.}
  \end{subfigure}\hfill
  \begin{subfigure}[t]{0.24\textwidth}
    \centering
    \includegraphics[width=\linewidth,height=5cm]{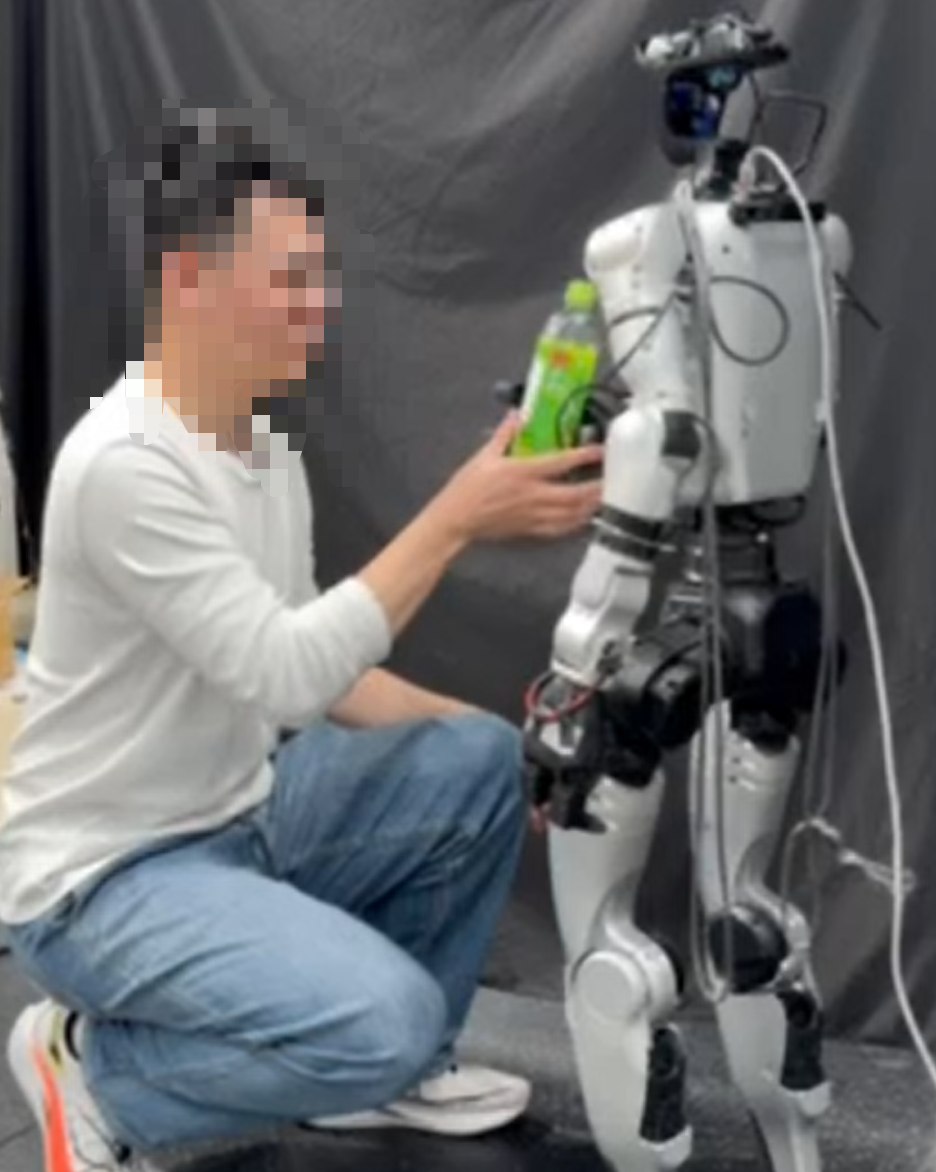}
    \caption{Handing over the drink.}
  \end{subfigure}\hfill
  \begin{subfigure}[t]{0.24\textwidth}
    \centering
    \includegraphics[width=\linewidth,height=5cm]{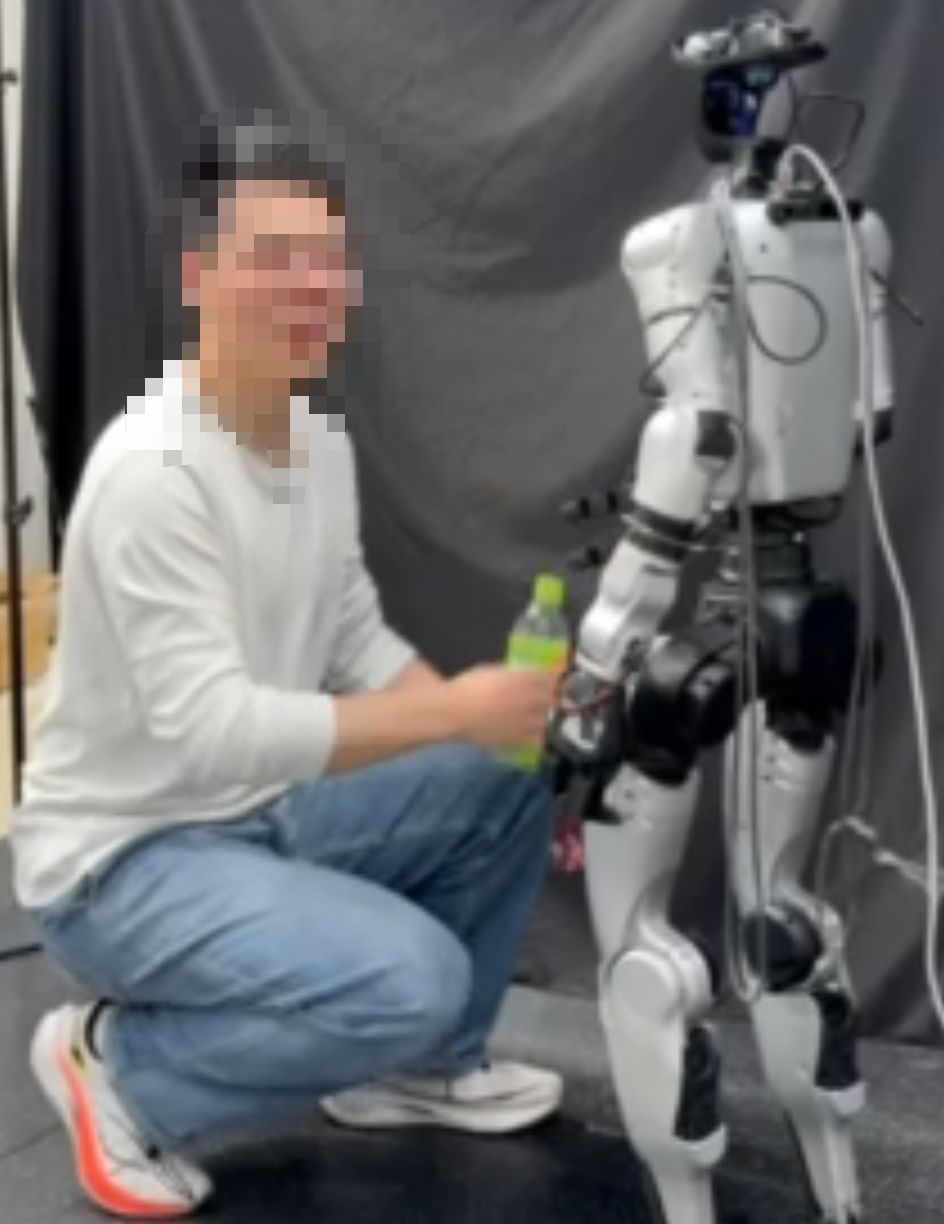}
    \caption{The user receives the drink.}
  \end{subfigure}
  \caption{Partial “Bring-me-a-drink” subtasks: the humanoid searches for the user, approaches them, hands over the drink, and completes the delivery.}
  \label{fig:52}
\end{figure*}

\subsection{Real-World Setup}\label{sec:setup}
We conduct real-world experiments on a Unitree G1 humanoid robot equipped with a Dex3-1 dexterous hand for manipulation and a head-mounted RGB-D camera for egocentric perception. 
Our system is deployed in a two-computer setup.
A desktop workstation with an RTX~5080 GPU runs the perception and decision modules, including SAM3-based multi-object segmentation, the VLM-based planning module, and the geometry-grounded supervisor.
The robot's onboard Unitree computer runs the low-level controllers, including the arm control stack and the RL locomotion policy.

Our experiments are performed in an indoor office environment that includes typical furniture and everyday objects.
The scene is cluttered to induce partial observability, occlusions, and viewpoint changes, providing a challenging testbed for loco-manipulation.

We evaluate the agent on a suite of long-horizon humanoid loco-manipulation tasks.
To enable direct comparison with Being-0, we include the same tasks \cite{being0}:
\texttt{Fetch-bottle}, \texttt{Deliver-box}, \texttt{Grasp-bottle}, \texttt{Place-box}, and \texttt{Place-coffee}.

Beyond these, we introduce additional long-horizon tasks not reported in Being-0 to stress open-ended instruction following and multi-object tabletop reasoning:
(i)  \texttt{Tidy-desk}, which requires repeatedly picking and placing multiple objects until the desk is cleared.
(ii)  \texttt{Tabletop-sorting}, which requires categorizing objects into different containers.
(iii)  \texttt{I'm in a bad mood, bring me a drink}, which requires fetching a drink and handing it over to the user.

\subsection{Results on Being-0-aligned Multi-step Tasks}
We report success rates on the Being-0-aligned multi-step task suite in Table~\ref{tab:being0_table1}.
We follow the task definitions and success criteria from Being-0 as closely as possible, and report our results over 10 trials per task.
Across multi-step routines, our agent achieves reliable completion by combining structured task programs with multi-object 3D grounding and predicate-based supervision.

\begin{table}[t]
\centering
\small
\setlength{\tabcolsep}{8pt}
\renewcommand{\arraystretch}{1.1}
\caption{Comparing the proposed method with being-0, the results show the number of successful manipulations in 10 trials.}
\label{tab:being0_table1}
\begin{tabular}{lcc}
\toprule
\textbf{Task} & {\footnotesize\textbf{Being-0}} & {\footnotesize\textbf{Cybo-Waiter}} \\
\midrule
\textbf{Fetch-bottle}   & 9/10 & 9/10 \\
\textbf{Deliver-basket} & 8/10 & 8/10 \\
\textbf{Grasp-bottle} & 8/10 & \textbf{10/10} \\
\textbf{Place-basket}    & 6/10 & \textbf{9/10} \\
\textbf{Place-coffee} & 6/10 & \textbf{8/10} \\
\bottomrule
\end{tabular}
\end{table}

\subsection{Results on Long-Horizon Tasks}\label{sec:lh_results}

Table~\ref{tab:table2} reports success counts over 10 trials for each task.
Our full system consistently outperforms the ablation \textbf{W/O Supervisor}, improving success from 5/10 to 7/10 on \texttt{Tidy-desk}, from 6/10 to 8/10 on \texttt{Tabletop-sorting}, and from 7/10 to 9/10 on \texttt{Bring-me-a-drink} (see Figs.~\ref{fig:51} and~\ref{fig:52} for qualitative examples).
These gains highlight the importance of the supervisor in long-horizon execution: by evaluating predicate-based preconditions and success conditions over stable frames, the supervisor reduces premature transitions caused by transient perception noise, detects off-nominal states, and triggers targeted recovery instead of restarting the entire task.
Overall, the results suggest that geometry-grounded monitoring and feedback are critical for maintaining reliable task progression when tasks require repeated object-level actions and extended navigation.

\begin{table}[t]
\centering
\small
\setlength{\tabcolsep}{8pt}
\renewcommand{\arraystretch}{1.1}
\caption{Success counts over 10 trials on additional long-horizon tasks. We compare the full system (Cybo-Waiter) with an ablation that disables the supervisor (W/O Supervisor).}
\label{tab:table2}
\begin{tabular}{lcc}
\toprule
\textbf{Task} & {\footnotesize \textbf{W/O Supervisor}} & {\footnotesize \textbf{Cybo-Waiter}} \\
\midrule
\textbf{Tidy-desk}   & 5/10 & 7/10 \\
\textbf{Tabletop-sorting} & 6/10 &\textbf{ 8/10} \\
\textbf{Bring-me-a-drink} & 7/10 & \textbf{9/10 }\\
\bottomrule
\end{tabular}
\end{table}

\section{Conclusion and Limitations}

We have presented a VLM-driven hierarchical agent framework for humanoid locomotion-manipulation task execution. By combining task-conditioned grounding, segmentation-grounded geometric monitoring, structured feedback-driven replanning, and a humanoid integrated execution layer, the proposed system has enabled more reliable long-horizon execution and recovery. Experiments on representative tasks have demonstrated the effectiveness of the framework and have highlighted the value of explicit monitoring and structured recovery in humanoid agent systems.

\begin{appendix} \label{appendix}

\noindent\textbf{Example.} A minimal plan snippet following our schema is:
\begin{lstlisting}[language={},caption={},label={}]
{
  "task_id":"6e0d3831",
  "command":"Help me tidy up the desk.",
  "subtasks":[
    {"subtask_id":1,
     "type":"MANIPULATION",
     "name":"subtask_1",
     "target":{"ref":"obj_1",
               "category":"unknown",
               "attributes":[],
               "relations":[],
               "phrase":"cup"},
     "destination":{"phrase":"container"},
     "manipulation":{"arm":"RIGHT",
                     "action":"PLACE",
                     "use_upper_body_mpc":true},
     "locomotion":{},
     "preconditions":[{"key":"VISIBLE",
                       "args":{"object":"obj_1"},
                       "op":"==",
                       "value":true,
                       "stable_frames":3}],
     "success_conditions":[{"key":"SUPPORTED_BY",
                            "args":{"object":"obj_1",
                            "support":"table_1"},
                            "op":"==",
                            "value":true,
                            "stable_frames":10}],
     "failure_handlers":[],
     "timeout_sec":30,
     "max_retry":2}
  ]
}
\end{lstlisting}
\end{appendix}

\vspace{12pt}

\bibliographystyle{IEEEtran}
\bibliography{references} 

\end{document}